\documentclass{article}

\usepackage{amssymb}
\usepackage{algorithm}
\usepackage{algorithmic}
\usepackage{amsmath}
\usepackage{caption}
\usepackage{graphicx}
\usepackage{wrapfig}
\usepackage{booktabs}
\usepackage{multirow}
\usepackage{xcolor}
\usepackage{float}
\usepackage{array}
\usepackage[none]{hyphenat}
\usepackage{enumitem}
\usepackage[breakable]{tcolorbox}
\usepackage{xcolor}
\usepackage{amssymb}
\usepackage{hyperref}

\usepackage[preprint]{corl_2025}

\title{VLA-Touch: Enhancing Vision-Language-Action Models with Dual-Level Tactile Feedback}

\author{
  Jianxin Bi$^{\star1}$,
  Kevin Yuchen Ma$^{1}$,
  Ce Hao$^1$,
  Mike Zheng Shou$^{1}$,
  Harold Soh$^{\star1,2}$ \\
  $^1$Dept. of Computer Science, National University of Singapore \\
  $^2$Smart Systems Institute, NUS\\
  \vspace{0.5em}
  $^{\star}$Correspondence: \{jianxin.bi, harold\}@comp.nus.edu.sg
}

\begin{document}
\maketitle

\vspace{-6mm}

\begingroup
\hyphenpenalty=50  
\exhyphenpenalty=50
\sloppy  

\begin{abstract}
Tactile feedback is generally recognized to be crucial for effective interaction with the physical world. However, state-of-the-art Vision-Language-Action (VLA) models often lack the ability to interpret and use tactile signals, limiting their effectiveness in contact-rich tasks. Incorporating tactile feedback into these systems is challenging due to the absence of large multi-modal datasets.
We present VLA-Touch, an approach that enhances generalist robot policies with tactile sensing \emph{without fine-tuning} the base VLA. Our method introduces two key innovations: (1) a pipeline that leverages a pretrained tactile-language model that provides semantic tactile feedback for high-level task planning, and (2) a diffusion-based controller that refines VLA-generated actions with tactile signals for contact-rich manipulation. Through real-world experiments, we demonstrate that our dual-level integration of tactile feedback improves task planning efficiency while enhancing execution precision. Code is open-sourced at \href{https://github.com/jxbi1010/VLA-Touch}{this URL}.

\end{abstract}

\endgroup


\section{Introduction} \label{Sec: introduction}

Recent advances in Vision-Language-Action (VLA) models \cite{ahn2022icanisay, kim2024openvla, rt22023corl, open_x_embodiment_rt_x_2023, liu2025rdtb, intelligence2025pi05visionlanguageactionmodelopenworld} have  improved general-purpose policy learning, enabling robots to interpret high-level instructions and execute tasks across diverse environments.
However, many real-world tasks—particularly those involving contact-rich interactions—remain challenging due to the inherent limitations of visual perception alone \citep{funk2025importancetactilesensingimitation}. Vision cannot reliably determine  object compliance, surface texture, or contact events due to visual ambiguity. While prior research has explored integrating tactile sensing into task-specific policies \citep{he2024foarforceawarereactivepolicy,xue2025reactive,ai2024robopacklearningtactileinformeddynamics}, incorporating tactile information into large-scale foundation models remains underexplored.


\begin{figure}[t]
    \centering
    \includegraphics[width=1.0\textwidth]{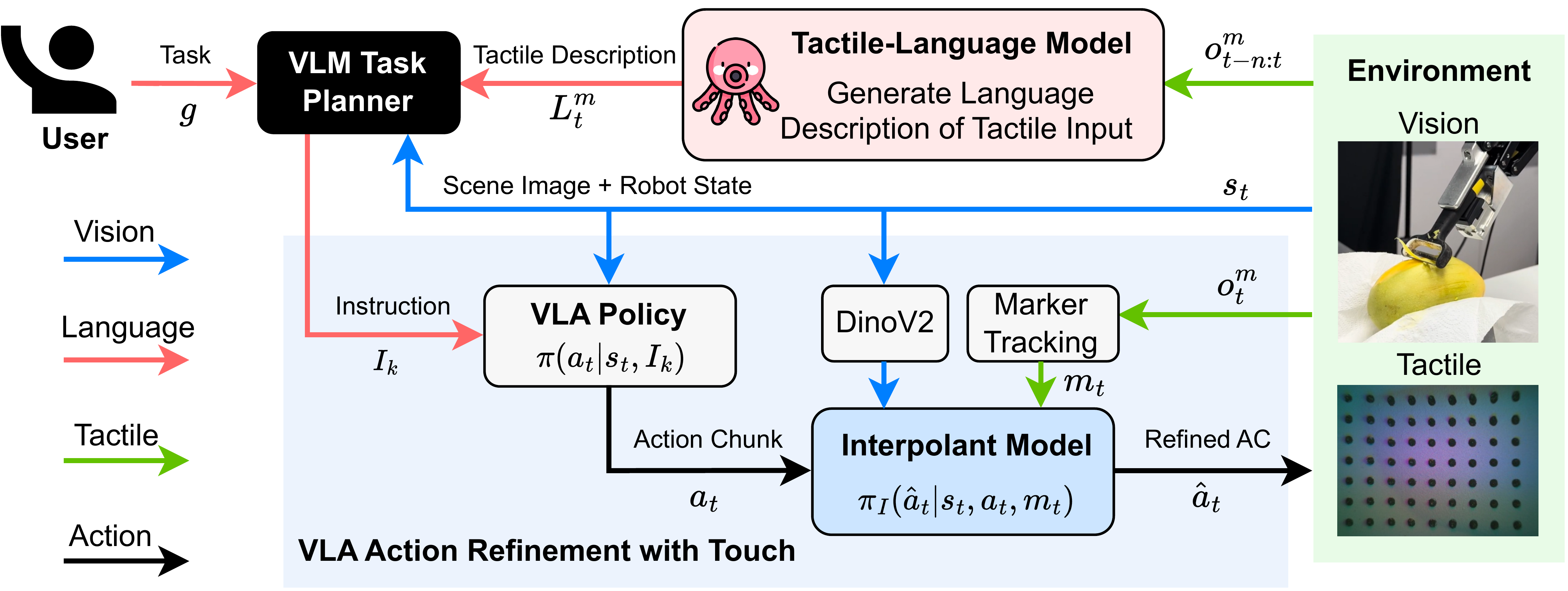}
    \caption{VLA-Touch incorporates Dual-level Tactile Feedback for Planning and Manipulation. \textbf{Planning}: Given a scene image $s_t$ and task goal $g$, the VLM Task Planner generates manipulation instruction $I_k$ for policy execution. A tactile-language model (Octopi) converts a sequence tactile input $o^m_{t-n:t}$ to language description $L^m_t$, which informs VLM for updated instruction. \textbf{Control}: The base VLA $\pi(a_t|s_t,I_k)$ generates action chunk $a_t$ based on visual observation $s_t$ and instruction $I_k$. The action chunk is then refined by an interpolant policy $\pi_I(\hat a_t|s_t,a_t,m_t)$ that takes as input both visual embeddings from a pretrained DinoV2 model and low-dimensional tactile signals $m_t$ processed a marker tracking algorithm from raw tactile input $o^m_t$.}
    \label{Fig: framework}
    \vspace{-10pt}
\end{figure}

We hypothesize that tactile feedback benefits contact-rich manipulation at two levels: 1) at the \textbf{planning} level, tactile feedback enables reasoning about object properties such as hardness and surface roughness that are visually ambiguous; and 2) at the \textbf{control} level, tactile sensing informs contact dynamics including surface friction and contact event detection for precise manipulation.
However, incorporating tactile sensing into large-scale foundation models presents challenges. Current VLA models are predominantly trained on visual, language, and action data~\cite{open_x_embodiment_rt_x_2023}, with no intrinsic mechanisms to process or reason over tactile inputs; this  fundamentally limits their ability to leverage tactile feedback for both planning and control in contact-rich scenarios. 
Developing tactile-informed embodied AI systems requires addressing two key challenges: 1) how to enable an agent to effectively acquire and plan with tactile information when most pretrained models have no prior exposure to tactile data, and 2) how to integrate tactile sensing into existing VLA pipelines when the base models lack dedicated tactile input modalities? 

In this work, we take a step toward addressing the above challenges. Specifically, we investigate how tactile signals can be incorporated into existing pipelines with \textit{minimal} modification. Our approach is modular: functional but loosely coupled, allowing us to probe the utility of touch in both planning and control without fine-tuning base VLA models. 
We propose a dual-level tactile feedback framework (Fig. \ref{Fig: framework}) inspired by the human neural system, consisting of:
\begin{enumerate}[noitemsep,topsep=0pt,leftmargin=2em]
\item A Task planner (analogous to higher cognitive functions in the prefrontal cortex \cite{miller2001integrative}): a Vision-Language Model that processes task goals, scene images, and linguistic feedback to reason about the task and generate manipulation instructions for VLA execution; 
\item A Tactile-Language Model (analogous to the Secondary Somatosensory Cortex \cite{tame2015early}): a pretrained model that converts tactile sensing data into linguistic descriptions of object properties such as hardness and texture, providing tactile feedback to the task planner for updated planning;
\item A Tactile-Augmented Controller (analogous to the Posterior Parietal Cortex \cite{culham2006human}): a controller trained to refine VLA-generated action sequences by incorporating tactile signals to achieve precise manipulation in contact-rich tasks.
\end{enumerate}

We conducted three real-world experiments to evaluate how dual-level tactile feedback affects task planning and control performance. Our experiments systematically examine: 1) how linguistic tactile feedback improves task planning in visually ambiguous scenarios, 2) how our tactile-enhanced controller refines VLA-generated actions for precise manipulation, and 3) the necessity of multi-modal sensor input through comprehensive ablation studies. Results demonstrate that our method of providing linguistic tactile feedback improves planning efficiency by up to 40\%, while our tactile-incorporated controller achieves up to 35\% higher manipulation success rates compared to vision-only VLA baselines. In addition, our dual-level tactile feedback system leads to up to 35\% higher task success rate than using tactile feedback only in planning or only in manipulation phase.
These findings support our hypothesis that dual-level tactile feedback is crucial for contact-rich manipulation. Additionally, our ablation studies show that multi-modal sensor fusion combining both vision and touch is important for achieving optimal performance.

To summarize, our work makes the following key contributions:
\begin{itemize}[noitemsep,topsep=0pt, leftmargin=2em]
\item A dual-level framework that enhances VLA models through tactile feedback, enabling both tactile-assisted task planning and policy refinement for contact-rich manipulation;

\item A diffusion-based controller that effectively refines VLA-actions through tactile sensing;

\item Comprehensive real-world experiments in contact-rich tasks, complemented by ablation studies that provide further understanding into the critical design elements of our framework.

\end{itemize}
Looking ahead, our work serves as a modular scaffold that enables  experimentation and isolation of tactile input effects, and represents an step toward tactile-aware embodied agents. The system shows clear benefits of tactile feedback, but there remains room for improvement. We plan to explore alternative forms of high-level and low-level tactile feedback, such as richer linguistic abstractions or continuous tactile embeddings. Tighter integration between the planner, tactile model, and controller may also reduce redundancy and latency. 
\section{Related Work} \label{Sec: related works}

\textbf{Vision-Language-Action Models.}
VLA models \citep{rt12022rss, octo_2023, rt22023corl, kim2024openvla, black2024pi0visionlanguageactionflowmodel, driess2023palmeembodiedmultimodallanguageicml, li2024visionlanguage, cheang2024gr2generativevideolanguageactionmodel, liu2025rdtb, intelligence2025pi05visionlanguageactionmodelopenworld, nvidia2025gr00tn1openfoundation} integrate visual perception with language understanding to generate robot actions for diverse tasks. These models typically leverage either pretrained Visual-Language Models (VLMs) \citep{driess2023palmeembodiedmultimodallanguage} or combine separate vision and language foundation models \citep{Zhai_2023_ICCV, oquab2023dinov2,chung2024scalinginstructionfinetunedlanguagemodels}, then train a unified backbone with an action decoder on robot manipulation datasets \citep{open_x_embodiment_rt_x_2023, fang2023rh20t}.
Current VLA models primarily rely on visual and proprioceptive feedback, which limits their effectiveness in contact-rich tasks where object properties and contact dynamics provide critical information. Recent work~\citep{jones2025sightfinetuninggeneralistrobot} has extended VLA architectures with additional sensory modalities, but focuses primarily on task-level reasoning rather than precise manipulation. \citep{hao2025tla} incorporates tactile feedback for language-conditioned policy but excludes visual perception.

\textbf{Tactile Foundation Models and Policy Learning.}
Tactile foundation models \citep{zhao2024transferable, higuera2024sparsh, feng2025learning} are pretrained on large-scale tactile datasets to extract generalizable representations from high-dimensional tactile signals across diverse sensor modalities \citep{gelsightmini, 9018215, 10474279, do2023densetact20opticaltactile}. Recent developments have extended these models into multi-modal frameworks that enable object reasoning with language \citep{yu2024octopi} and integrate vision and language for unified representation learning \citep{fu2024a}.
Our approach utilizes the tactile-language model from \citep{yu2024octopi} to generate semantic interpretations of grasped objects for task planning when visual information is ambiguous.
Very recent tactile-augmented policies have demonstrated improved manipulation capabilities through various methods—including augmented 3D tactile point observations \citep{huang20243dvitac}, tactile-informed dynamics models \citep{ai2024robopacklearningtactileinformeddynamics}, slow-fast architectures for reactive tactile behavior \citep{he2024foarforceawarereactivepolicy, xue2025reactivediffusionpolicyslowfast}, and robust tactile sensing \citep{zhao2025polytouchrobustmultimodaltactile}—these approaches primarily address control-level improvements for specific tasks.
Different from these prior works, our framework incorporates tactile feedback at two complementary levels, which we posit will improve tactile-informed task planning performance and manipulation precision within a unified system.

\textbf{Tactile Sensors.}
Tactile sensors provide contact information that enables dexterous manipulation and precise object interaction. Two primary types exist: (1) taxel-based sensors \citep{uskin2024,tacniq_tac02_2024} that use sensing element arrays to measure contact properties at discrete locations, and (2) vision-based sensors \citep{gelsightmini} that capture tactile information through optical changes in deformable surfaces.
These sensors measure contact forces, surface textures, and material properties that are difficult to obtain through vision alone. For a comprehensive overview of tactile sensing technologies, see \citep{li2020review}.
In this work, we use the GelSight Mini tactile sensor \citep{gelsightmini}, which features an elastomeric gel membrane with embedded $7\times9$ marker arrays and an internal camera. When objects contact the gel surface, deformation is recorded as RGB images from which surface geometry and contact force information can be extracted.

\section{Methodology} \label{Sec: method}
In this section, we present our approach for integrating tactile sensing to improve both high-level task planning and low-level action refinement. Our implementation uses Robot Diffusion Transformer (RDT) \cite{liu2024rdt} as the base VLA model, though our methodology remains applicable across various VLA architectures.


\subsection{Tactile-Assisted Task Planning}

Motivated by the understanding that tactile sensing provides localized perceptual information, we propose a cyclical task planning scheme to acquire tactile information and incorporate it into the planning process.
Our framework (Fig. \ref{Fig: framework}) consists of several integrated modules. A language-conditioned VLM task planner (GPT-4o in this work) generates specific manipulation instructions based on the given goal and current observations. A VLA model (RDT in our case) then generates action chunks to execute the given manipulation instruction. Upon completion of an instruction or exceeding the maximum execution steps, a tactile-language model converts the recent tactile signals to a linguistic tactile description, which is given to the planner to generate a new manipulation instruction until the task goal is achieved. This interleaved approach of high-level planning and low-level control follows established patterns in VLA architectures \cite{black2024pi0visionlanguageactionflowmodel, intelligence2025pi05visionlanguageactionmodelopenworld}.

The VLM is prompted with specific instructions defining its capabilities, response format, and interaction protocol.  Please refer to the Appendix \ref{app:prompt} for prompt details. At a high-level, the prompt establishes the following key elements:
\begin{enumerate}[noitemsep,topsep=0pt,leftmargin=2em]
\item \textbf{System capabilities:} The robot has a single arm with a gripper and a tactile sensor capable of classifying physical properties (hardness, roughness) and surface patterns.
\item \textbf{Response format:} For each step, the planner must provide: a) A primitive action described in a single sentence, involving one elemental robot action interacting with at most one object. b) Information needed to be retrieved (if applicable)
\item \textbf{Interaction protocol:} The planner is informed that after each action, it will receive feedback about information retrieved or action execution results, which it must use to plan the next action.
\end{enumerate}

For tactile-based object property inference, we employ Octopi \citep{yu2024octopi}, a pretrained tactile-language model developed for the Gelsight sensor. Octopi infers properties of contacted objects, including roughness and hardness, by processing a sequence of tactile signals (6 frames) from the Gelsight sensor. Once Octopi is triggered, the tactile signals are converted into linguistic feedback, used for subsequent task planning.

\subsection{VLA Policy Refinement with Tactile Sensing}

Since VLA models do not natively incorporate tactile signals, we adopt an interpolant-based diffusion controller called BRIDGeR~\cite{chen2024behavioral} to refine VLA-generated actions using tactile feedback. Unlike conventional diffusion models that start from Gaussian noise, BRIDGeR employs stochastic interpolants to diffuse from an informative source distribution (in our case, the VLA-generated action distribution).

Our interpolant controller, denoted $\pi_{I}(\hat a_{t:t+T_i}| a_{t:t+T_i}, s_t, m_t)$, takes as input a source action chunk $a_{t:t+T_i}$ of length $T_i \leq T_a$ (where $T_a$ is the full VLA-generated horizon) and conditions the refinement process on the current state $s_t$ and tactile signal $m_t$. The state $s_t$ includes RGB images and robot proprioception, while $m_t$ encodes the tactile force signal. 
The tactile force signal $m_t$ is estimated from marker displacements using a 7$\times$9 marker array, we compute per-marker force vectors, then aggregate them to obtain a summed force vector and magnitude $m_t = (X, Y, M)$.  This approach follows the methodology of \cite{xue2025reactive}, but uses only the aggregated force rather than individual marker forces to reduce noise and provide a more compact representation.
The controller outputs a refined action sequence $\hat{a}_{t:t+T_i}$ for execution.


\begin{algorithm}
  \caption{Tactile-Assisted Task Planning and Action Refinement}
  \label{alg:tac-plan-refine}
  \begin{algorithmic}[1]
    \REQUIRE Task goal $g$, Scene image $o_t$, VLA Observation $s_t$, Task planner $\text{GPT-4o}$, VLA model $\text{RDT}$, Tactile-Language Model $\text{Octopi}$, Interpolant Controller $\pi_{I}$
    \STATE Initialize instruction $I_0$ from task goal $g$ and scene image $o_t$ by GPT-4o, $k=0$, $t= 0$
    \STATE 
    \WHILE{task not completed}
        \STATE $s_t, m_t \gets \text{UpdateObs()}$ \COMMENT{Update visual observation, proprioceptive state, and tactile signal}
                
        \STATE \textbf{// Refine Action Chunk Segments and Execute}
        \WHILE{instruction $I_k$ not completed}
            \STATE $a_{t:t+T_a} \gets \pi(s_t, I_k)$ \COMMENT{Generate action chunks from RDT}
            \WHILE{$a_{t:t+T_a}$ not fully refined}
                \STATE $\hat{a}_{t:t+T_r} \gets \pi_I(s_t, a_{t:t+T_r}, m_t)$ \COMMENT{Refine $T_r$ steps of actions with tactile feedback}
                \STATE \text{Execute} refined action $\hat{a}_{t:t+T_r}$
                \STATE $t \leftarrow t+T_r$
                \STATE $s_t, m_t \gets \text{UpdateObs()}$
            \ENDWHILE
            \IF{execution failed}
                \STATE \textbf{break} \COMMENT{Exit execution loop for replanning}
            \ENDIF
        \ENDWHILE
        
        \STATE \textbf{// Update Instruction by Task Planner}
        \STATE $o^m_t \gets \text{GetTactileObs()}$ \COMMENT{Acquire tactile perception}
        \STATE $L_{m_t} \gets \text{Octopi}(o^m_t)$ \COMMENT{Convert tactile input to linguistic description}
        \STATE $I_{k+1} \gets \text{GPT-4o}(g, o_t, L_{m_t}, I_k)$ \COMMENT{Update instruction based on task progress}

        \STATE $k \gets k+1$
    \ENDWHILE
  \end{algorithmic}
\end{algorithm}

We train the controller via supervised learning on a dataset of paired VLA-generated and expert action sequences. This dataset is constructed by sampling states from expert demonstrations and generating corresponding source actions from the VLA model for Interpolant model to refine.
At inference time, we implement a sliding window refinement strategy that processes non-overlapping segments of the source action chunk sequentially. After executing a refined window, the controller advances to the next segment, starting from the endpoint of the previous cycle. This continues until the entire VLA action chunk has been refined and executed. Please see Algorithm~\ref{alg:tac-plan-refine} for an overview and Appendix \ref{app:interpolant} for implementation details.

\section{Experiments} \label{Sec: experiments}

In this section, we describe results from real-world experiments designed to evaluate our main hypothesis that dual-level feedback is crucial for contact-rich manipulation tasks. Specifically, we seek to answer the following research questions: 

\begin{enumerate}[noitemsep,topsep=0pt,leftmargin=3em]
\item[\textbf{Q1.}] Does tactile feedback improve task planning performance? Are linguistic descriptions of tactile information sufficient for VLM-based planning (versus raw tactile images)?

\item[\textbf{Q2.}] Does tactile feedback benefit VLA policies for control? In addition, how does the interpolant-based diffusion compare against simple residual controller?

\item[\textbf{Q3.}] How does our dual-level tactile feedback system benefit performance in contact-rich tasks? What if we have only high-level or low-level feedback? 
\end{enumerate}

\begin{figure} [t]
    \setlength{\belowcaptionskip}{-6pt}
    \centering
    \includegraphics[width=0.99\textwidth]{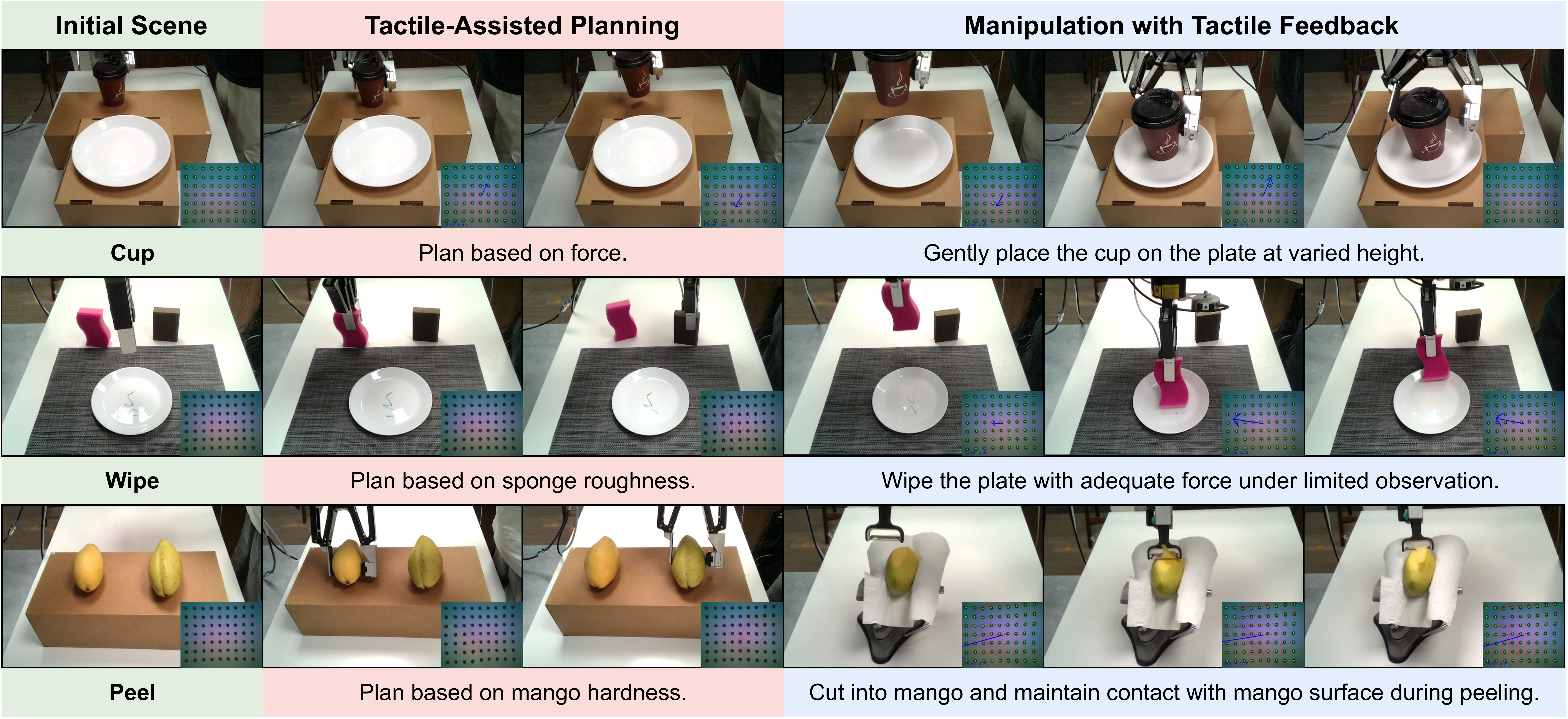}
    \caption{Experiments pipeline with images captured by the scene camera. Each task contains two stages: \textbf{Planning}: inference over contact properties (force, roughness, hardness) with tactile perception for tactile-informed task planning. \textbf{Manipulation}: refine action with tactile feedback for contact-rich manipulation.
    } 
    \label{Fig: exp showing}
\end{figure}

\vspace{-10pt}
\subsection{Experiment Setup}
\textbf{Hardware Setup}.
We use a Franka Emika Panda manipulator equipped with a Robotiq 2F-140 gripper. A GelSight Mini tactile sensor is mounted on one finger of the gripper. Visual perception is provided by two RealSense cameras: one fixed overhead as a scene camera and one mounted on the robot's wrist. All sensor streams are processed on a workstation with an RTX 4090 GPU for inference.

\textbf{Baseline methods.} We evaluate our approach against several baseline methods across both task planning and manipulation components. To answer \textbf{Q1}, we test whether tactile feedback improves planning, and whether structured (linguistic) representations outperform raw tactile images when provided to a off-the-shelf VLM.
We compare three ways of using GPT-4o for task planning: 
\begin{itemize}[noitemsep,topsep=0pt,leftmargin=2em]
\item \textbf{GPT-4o:} Receives only the scene image (RGB from the fixed camera); no tactile input.
\item \textbf{GPT-4o + Tactile Image:} Receives the scene image and a sequence of raw tactile images from the GelSight sensor.
\item \textbf{GPT-4o + Octopi (Ours):} Receives the scene image and linguistic tactile feedback, generated by converting tactile images into textual descriptions using Octopi. 
\end{itemize}

To answer \textbf{Q2} regarding manipulation, we evaluate the effectiveness of tactile feedback for contact-rich manipulation and assesses which controller architecture better integrates touch for policy refinement. We use Robot Diffusion Transformer (RDT-1B) as our base VLA policy, and compare RDT against two tactile-enhanced variants:
\begin{itemize}[noitemsep,topsep=0pt,leftmargin=2em]
    \item \textbf{RDT:} Vanilla RDT without tactile feedback.
    
    \item \textbf{RDT + Residual Controller:} An RNN-based controller that incorporates tactile signals to refine RDT action chunk by predicting residual actions.
    
    \item \textbf{RDT + Interpolant Controller:} The Interpolant controller that refines RDT action chunk conditioned on tactile signals.

\end{itemize}

\textbf{Dual-level Tactile Feedback.} To answer \textbf{Q3}, we ablate our system to compare against two single-level variants: one with tactile feedback used only during planning, and one with tactile feedback used only during manipulation. This isolates the contribution of high-level vs. low-level tactile feedback.

\begin{itemize}[noitemsep,topsep=0pt,leftmargin=2em]
    \item \textbf{w/o Planning: } Planning with GPT-4o, manipulation with RDT + Interpolant controller.
    
    \item \textbf{w/o Control:} Planning with GPT-4o + Octopi, manipulation with RDT.
    
    \item \textbf{VLA-Touch:} Planning with GPT-4o + Octopi, manipulation with RDT + Interpolant controller.

\end{itemize}

\textbf{Tasks.} We evaluate on three contact-rich tasks, each requiring tactile inference for both high-level planning and low-level control:
\begin{itemize}[noitemsep,topsep=0pt,leftmargin=2em]
\item \textbf{Cup:} The robot must pick up a lidded cup, determine (through tactile feedback) whether it contains water, and place it at corresponding location without spilling.
\item \textbf{Wipe:} The robot selects the smoother of two sponges using touch, then wipes ink off a plate.
\item \textbf{Peel:} The robot identifies the softer or harder of two mangoes via touch and peels the selected one with a hand-held peeler.
\end{itemize}

\vspace{-5pt}
\subsection{Task Planning Results}

\begin{figure} [t]
    \setlength{\belowcaptionskip}{-4pt}
    \centering
    \includegraphics[width=1.0\textwidth]{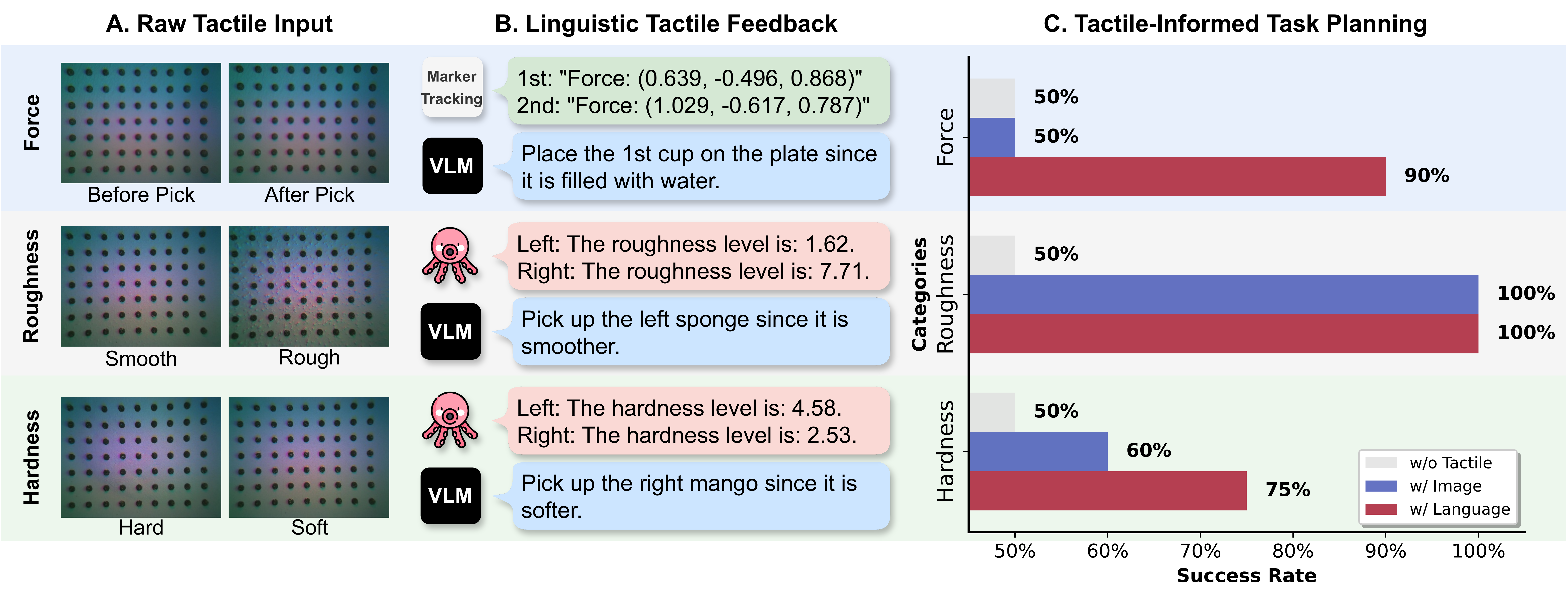}
    \caption{
    \textbf{A}: Raw tactile images from three tasks requiring feedback on force, surface roughness, and hardness.
\textbf{B}: Corresponding linguistic tactile descriptions generated by the Tactile-Language Model (Octopi) and GPT-4o’s responses.
\textbf{C}: Task planning outcomes under three conditions: GPT-4o with no tactile input, with raw tactile images, and with linguistic tactile descriptions. Each condition was evaluated over 20 trials using tactile data collected from grasps executed by the VLA model with the Interpolant Controller.}
    \label{Fig: table of main results}
    
\end{figure}

Our experimental results show that tactile feedback significantly improves task planning performance, and linguistic tactile descriptions are more effective than raw tactile images for VLM-based planning.
As shown in Figure~\ref{Fig: table of main results}C, GPT-4o fails to infer over tactile properties when given only the scene image; its responses are consistently non-committal (e.g., ``I can’t determine''), and forced choices result in success rates comparable to random guessing. Providing raw tactile images enables GPT-4o to identify surface roughness effectively (100\% success), but it struggles with interpreting force (50\%) and object hardness (60\%). In contrast, when supplied with linguistic tactile descriptions generated by Octopi, performance improves substantially—achieving 90\% success for force prediction and 75\% for hardness. These results indicate that tactile feedback is essential for planning in contact-rich tasks, and that structured, language-based representations are more usable by GPT-4o than raw tactile inputs. Refer to Appendix \ref{app:ablation} for extra study.

For force evaluation specifically, we implemented a separate approach since Octopi is not pretrained for force estimation. We use the marker tracking algorithm to get estimated force vectors and provide to GPT-4o along with baseline reference vectors. The reference values serve as calibration points when force estimates are relative or unnormalized, though they are less critical when methods report forces in absolute units (e.g., Newtons).

\subsection{Manipulation Results}

\begin{table*}[ht]
    \centering
    \caption{Manipulation performance across different tasks}
    \resizebox{0.8\textwidth}{!}
    {
    \begin{tabular}{l|l|ccc|cc}
        \toprule
        \textbf{Tasks}&\textbf{Evaluations}& \textbf{RDT} & \textbf{Residual} & \textbf{Interpolant} & \textbf{w/o Touch} & \textbf{w/o Vision}\\
        \midrule
        \multirow{2}{*}{\textbf{Cup}} 
        &\textbf{Pick}  & $9/20$ &$7/20$  & $\mathbf{12/20}$ & $10/20$ & $10/20$\\
        &\textbf{Place}  & $7/20$ &$6/20$  &$\mathbf{10/20}$  & $5/20$& $7/20$\\
        \midrule
        \multirow{2}{*}{\textbf{Wipe}} 
        &\textbf{Pick}  & $11/20$ &$15/20$  &$\mathbf{17/20}$ & $15/20$& $15/20$\\
        &\textbf{Wipe Partial}  & $8/20$ &$13/20$  & $\mathbf{16/20}$ & $12/20$ &$10/20$\\
        &\textbf{Wipe}  & $5/20$ &$6/20$  &$\mathbf{12/20}$ & $7/20$ &$8/20$\\
        \midrule
        \multirow{2}{*}{\textbf{Peel}} 
        &\textbf{Pick}  & $13/20$ &$14/20$  &$\mathbf{18/20}$ &$16/20$ & $13/20$\\
        &\textbf{Peel Partial}  & $8/20$ &$12/20$  &$\mathbf{13/20}$ &$12/20$  & $8/20$ \\
        &\textbf{Peel}  & $6/20$ &$7/20$  &$\mathbf{10/20}$  &$5/20$  & $5/20$\\

        \bottomrule
    \end{tabular}
    }
    \label{tab:main}
    
\end{table*}

Our experimental results demonstrate that tactile feedback significantly enhances performance in contact-rich manipulation tasks, and the interpolant-based controller (ours) outperforms both the base VLA policy and LSTM-based residual controller. As shown in Table~\ref{tab:main}, our interpolant controller consistently achieves the highest success rates across all tasks and evaluation metrics. The interpolant controller improves base RDT task success rates by $42\%$ (Cup), $140\%$ (Wipe), and $67\%$ (Peel) respectively; it also outperforms the residual controller method, achieving $67\%$ (Cup), $100\%$ (Wipe), and $42\%$ (Peel) higher task success rate. These results support the notion that tactile feedback substantially benefits VLA policies for manipulation control, and that interpolant-based diffusion controllers outperform simple residual controllers by better capturing the multi-modal nature of demonstration data.

\textbf{Cup.} The cup task reveals critical differences in grasping precision and contact awareness. RDT generates imprecise trajectories that cause the gripper to push the cup forward before closing, resulting in low pick success ($9/20$) and premature releases during placement due to insufficient tactile feedback ($2/9$ successful picks). The residual controller exhibits systematic positioning errors, with refined actions consistently posterior to desired positions ($7/20$), leading to unstable grasps and water spillage ($3/20$). This behavior reflects the controller's inability to handle the multi-modal nature of cup positioning. In contrast, our interpolant controller effectively corrects RDT's grasping behavior, achieving $12/20$ successful picks and demonstrating superior contact awareness with only one premature release during placement ($1/12$ successful picks).

\textbf{Wipe.} Successful wiping requires grasping the sponge at its center. RDT occasionally generates incorrect grasping trajectories (pushing sponge downward or grasping the sponge's edge), leading to incomplete wiping in $5/11$ successful picks.
The residual controller improves grasping but struggles with peripheral wiping, achieving only partial cleaning in $13/15$ trials, primarily in central regions, which indicates a limitation in capturing multi-modal action distributions.
Our interpolant controller was better at both grasping and pressure modulation, successfully completing full wiping in $12/20$ trials.

\begin{wrapfigure}{r}{0.30\textwidth}
    \vspace{-5pt}
    \setlength{\belowcaptionskip}{-30pt}
    \centering
    \includegraphics[width=\linewidth]{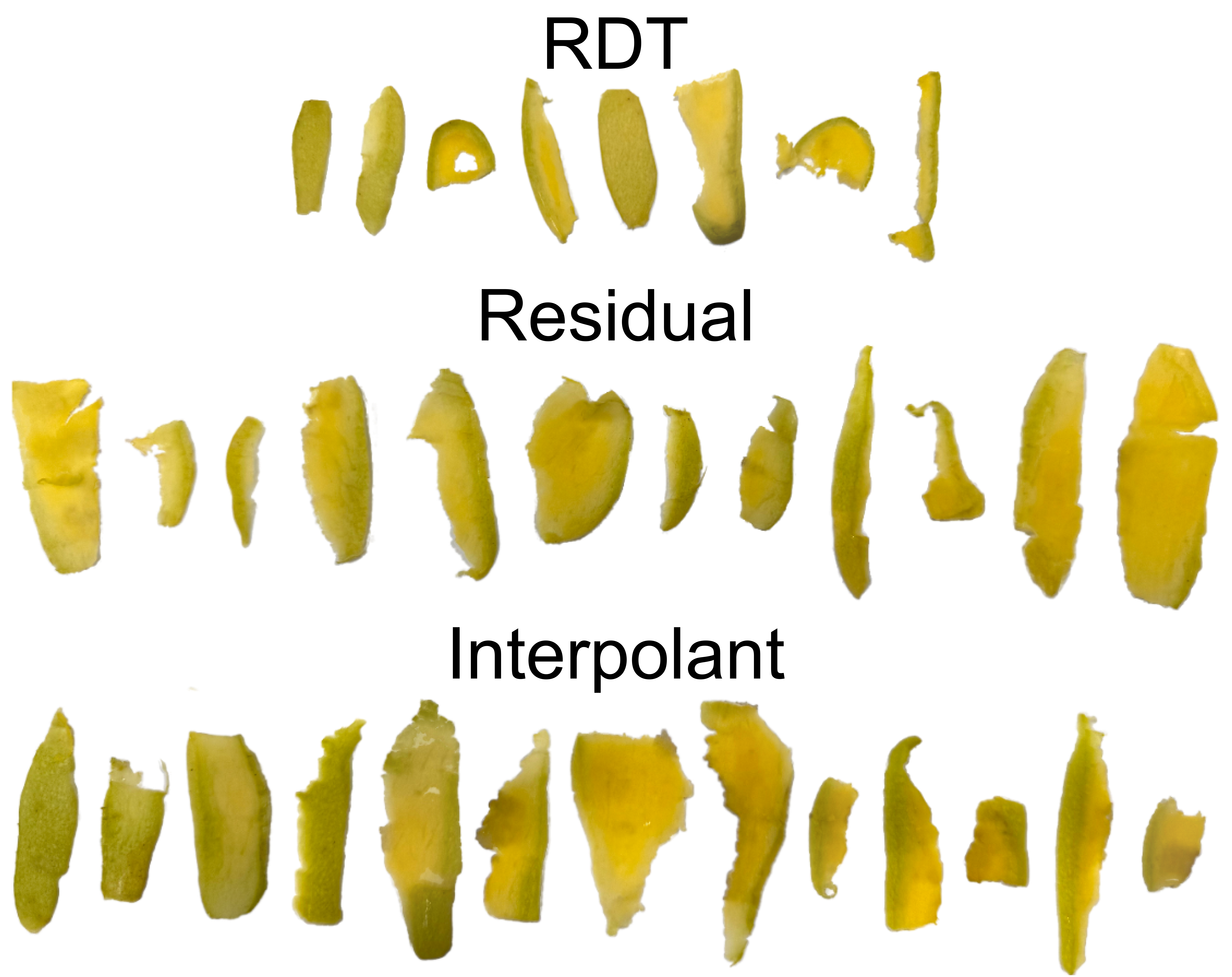}
    \caption{20 trails of peeling results, demonstrating the qualitative effect of tactile feedback on peeling.}
    \label{fig:peeling_res}
\end{wrapfigure}

\textbf{Peel.} 
The mango peeling task requires contact positioning and sustained pressure throughout the motion. RDT frequently slides without cutting (5/13 successful grasps) due to the lack of tactile feedback. The residual controller shows improved cutting initiation but prematurely elevates the peeler midway through ($5/14$ successful grasps), likely due to learning averaged policies that terminate early on larger mangoes. Interpolant controller maintains better contact throughout the process, as evidenced in Figure~\ref{fig:peeling_res}: while RDT removes minimal surface material and the residual controller produces 12 pieces with 5 short fragments due to contact loss, Interpolant controller removes 13 pieces with only 3 short fragments, demonstrating improved contact maintenance and cutting consistency.

\subsection{Dual-Level Tactile Feedback}

Our experimental results demonstrate that dual-level tactile feedback is essential for optimal performance, with either component alone leading to substantial degradation. As shown in Table~\ref{tab:dual_level_results}, removing tactile feedback from planning reduces success rates by $44$\% (Cup), $58$\% (Wipe), and $14$\% (Peel),  while removing it from control decreases performance by $33\%$ (Cup), $58\%$ (Wipe), and $43\%$ (Peel). 
Tactile-informed planning enables task-level decisions impossible with vision alone, such as distinguishing between empty and filled cups or selecting the smoother sponge for more effective wiping. For manipulation, tactile feedback provides contact information essential for pressure modulation, contact maintenance, and preventing premature releases. These results and observations confirm that both planning and control components critically depend on tactile information.
The ablated systems exhibit failures consistent with individual component evaluations: incorrect object selections without planning-level tactile feedback; failed grasping or inaccurate contact control due to the lack of control-level tactile feedback.

\begin{table*}[ht]
    \centering
    \caption{Interleaved Plan and Manipulation Evaluation.}
    \resizebox{0.55\textwidth}{!}
    {
    \begin{tabular}{l|ccc}
        \toprule
        \textbf{Tasks} &\textbf{w/o Planning} & \textbf{w/o Control} & \textbf{VLA-Touch}\\
        \midrule
        {\textbf{Cup}} & $5/20$ &$6/20$  &$\mathbf{9/20}$\\
        {\textbf{Wipe}} & $5/20$ &$5/20$  &$\mathbf{12/20}$\\
        {\textbf{Peel}} & $6/20$ &$4/20$  &$\mathbf{7/20}$\\
        \bottomrule
    \end{tabular}
    }
    \label{tab:dual_level_results}
\end{table*}

\vspace{-10pt}
\subsection{Ablation Study of Interpolant Controller}
We examine the necessity of multi-modal feedback for the Interpolant controller through ablation. Our experimental results demonstrate that both tactile and visual feedback are essential for the Interpolant Controller, with tactile feedback being particularly critical for contact-rich manipulation phases and visual feedback being important for spatial positioning and grasp planning. As shown in Table~\ref{tab:main}, removing tactile feedback ("w/o Touch") significantly degrades performance in contact-intensive tasks, reducing cup placement success by $50\%$, wiping by $42\%$, and peeling by $50\%$. Removing visual feedback ("w/o Vision") shows less severe but notable degradation, particularly affecting cup placement ($30\%$ reduction) and task success across all scenarios.

\textbf{Tactile Feedback.} The removal of tactile perception maintains comparable performance for basic picking tasks but severely impacts contact-rich manipulation phases. Without tactile feedback, the controller cannot detect contact forces, friction coefficients, or tool-object interactions, leading to incomplete task executions despite preserved manipulation capabilities: cup placement success rate drops by $50\%$, wiping drops by $42\%$, and peeling drops by $50\%$.

\textbf{Visual Feedback.} While the Interpolant Controller maintains reasonable performance without visual feedback, degradation is evident in spatial positioning tasks. For wiping, pick success rates remain acceptable ($15/20$), but the absence of visual information leads to suboptimal grasping positions (often at sponge edges), subsequently impairing effective plate cleaning (complete wiping drops from $12/20$ to $8/20$). Similarly, for peeling tasks, visual feedback loss affects cutting precision and positioning, contributing to reduced complete peeling success ($10/20$ to $5/20$).

\section{Conclusion} \label{Sec: conclusion}
\vspace{-10pt}
In this work, we presented VLA-Touch, a novel framework to improve VLA policies with dual-level tactile feedback without fine-tuning the VLA model with tactile data. 
Our work introduces two key innovations: (1) a pipeline leveraging a pretrained tactile-language model for semantic tactile feedback in high-level task planning, and (2) a interpolant-based controller that refines VLA-generated actions with tactile signals for contact-rich manipulation. 
Through three real-world experiments, we demonstrated the effectiveness of our dual-level tactile feedback system.
Ablation studies confirmed that both visual and tactile are crucial for contact-rich tasks. Our work takes one step toward more capable embodied agents that can leverage multiple sensory modalities for physical interactions in the real world.

\section{Limitation and Future Works}
Our current implementation has several limitations. First, the gripper control setup differs from the Octopi pretraining dataset; this creates discrepancies in contact measurements that can compromise tactile interpretation. We observed this particularly for object hardness. 
Second, we mainly focussed on generalization across object positioning and target locations in our experiments (varying cup placement heights, different wiping ink patterns, and diverse mango specimens); future work could examine generalization across tasks but this would require significantly more training. 
Third, the Interpolant controller operates at 8 hz and does not yet fully leverage high-frequency tactile signals ($\geq$ 25 Hz) that could enhance performance during dynamic interactions. 

We identify three directions for future research: (1) developing an active grasping framework that dynamically refines grasp poses through combined visual-tactile feedback, enabling more sophisticated touch inference; (2) designing an event-triggered inference strategy that accelerates processing during contact state transitions while conserving resources during stable periods; and 3) developing task-agnostic policy refinement method that can scale to more real-world tasks.
These advancements would address limitations in grasp quality and temporal resolution while expanding the capabilities of tactile-enhanced manipulation.

\clearpage
\bibliography{references}

\clearpage
\appendix
\appendix
\setcounter{section}{0}

\tcbset{
  onecol/.style={
    colback=gray!10,
    colframe=gray!80,
    boxrule=0.5pt,
    arc=2mm,
    fontupper=\scriptsize\ttfamily,
    left=1mm,
    right=1mm,
    top=1mm,
    bottom=1mm,
    boxsep=1mm,
    breakable,
    width=\linewidth,
  }
}

\tcbset{
  twocol/.style={
    colback=gray!10,
    colframe=gray!80,
    boxrule=0.5pt,
    arc=2mm,
    fontupper=\scriptsize\ttfamily,
    left=1mm,
    right=1mm,
    top=1mm,
    bottom=1mm,
    boxsep=1mm,
    nobeforeafter,
    breakable,
    width=(\linewidth-2mm)/2,
  }
}

\part*{Appendix}

\section{Prompt Examples}
\label{app:prompt}

For all tasks, we first give the VLM the following global context:

\begin{tcolorbox}[onecol]
System: You are a robot highly skilled in robotic task planning and interactive reasoning, adept at planning actions to retrieve unknown information for reasoning and decision making, and subsequently breaking down intricate and long-term tasks into distinct primitive actions. As a robot, you have one arm with a gripper. You have a tactile sensor mounted on the gripper, which can be used to classify physical properties (hardness, roughness, weight) of objects. Your task is to plan out steps of actions to take to retrieve information and complete the task. For each time, return 1. a primitive action in the form of one sentence, it should contain one elemental robot action interacting with at most one object 2. information needed to retrieve if applicable. After each action, feedback will be given back to you for information retrieval or action execution, you will then plan the next robot action based on the feedback. The feedback can be in the form of property classification results or raw tactile sensor images. Remember to refer to objects by their spatial locations (like left or right). Only give physical actions that the robot has to execute. Keep the action and information needed concise. Only give one action step in each response and wait for the user feedback.

\end{tcolorbox}

Each task is then initiated by a task planning request. In our experiments, we compared two ways to provide tactile feedback: a. to provide raw tactile images; and b. to provide linguistic tactile feedback generated by Octopi and image processing (ours). Examples of the two methods are given below.

\textbf{Task 1: Cup}

\begin{tcolorbox}[onecol, adjusted title=Raw images]
User: There is a cup in the image. Step by step, identify whether the cup is full or empty. If it is not empty, put it onto the plate.
\vspace{2mm}

Robot: 1. **Action:** Grip the cup gently to lift it slightly. \\
- **Information Needed:** Determine the weight of the cup to assess if it is full or empty.
\vspace{2mm}

User: The first image is the gelsight tactile sensor image before touching the object: <image1/>. \\The second image is the gelsight tactile sensor image after lifting the object: <image2/>
\vspace{2mm}

Robot: The tactile sensor images indicate that the cup is likely empty, as there is minimal change in the sensor pattern. \\2. **Action:** Place the cup back in its original position.
\end{tcolorbox}

\begin{tcolorbox}[onecol, adjusted title=Ours]
User: There is a cup in the image. Step by step, identify whether the cup is full or empty. If it is not empty, put it onto the plate.

\vspace{2mm}
Robot: 1. Action: Grip the cup gently to lift it slightly. \\- Information Needed: Determine the weight of the cup to assess if it is full or empty.

\vspace{2mm}
User: The tactile shear force vector has magnitude: 0.6571505252217286, xy-direction: [-0.6038576879954624, 0.7970921481540104]. \\For reference, the max force magnitude is around 0.55 for an empty cup and around 1.1 for a full cup.

\vspace{2mm}
Robot: 2. Action: Place the cup onto the plate. \\Information Needed: None.
\end{tcolorbox}

\textbf{Task 2: Wipe}

\begin{tcolorbox}[onecol, adjusted title=Raw images]

User: There are two sponges in the image, one on the left and another on the right. Step by step, touch and identify the roughness of the sponges, and pick up the smoother one.

\vspace{2mm}

Robot: 1. Touch the sponge on the left to assess its roughness. \\Information needed: Roughness level of the left sponge.

\vspace{2mm}

User: The first image is the gelsight tactile sensor image before touching the object: <image1/>. \\The second image is the gelsight tactile sensor image after touching the object:  <image2/>

\vspace{2mm}

Robot: The tactile sensor images show minimal deformation, indicating that the left sponge is relatively smooth. \\Next, touch the sponge on the right to assess its roughness. \\Information needed: Roughness level of the right sponge.

\vspace{2mm}

User: The first image is the gelsight tactile sensor image before touching the object: <image1/>. \\The second image is the gelsight tactile sensor image after touching the object:  <image2/>

\vspace{2mm}

Robot: The tactile sensor images show significant deformation, indicating that the right sponge is rougher than the left sponge. \\Next, pick up the sponge on the left, as it is smoother.
\end{tcolorbox}

\begin{tcolorbox}[onecol, adjusted title=Ours]

User: There are two sponges in the image, one on the left and another on the right. Step by step, touch and identify the roughness of the sponges, and pick up the smoother one.

\vspace{2mm}

Robot: Touch the sponge on the left to assess its roughness.

\vspace{2mm}

User: The hardness level is: 2.973193645477295, The roughness level is: 6.088855266571045

\vspace{2mm}

Robot: Touch the sponge on the right to assess its roughness.

\vspace{2mm}

User: The hardness level is: 4.499370098114014, The roughness level is: 7.802529811859131

\vspace{2mm}

Assistant: Pick up the sponge on the left, as it is smoother.
\end{tcolorbox}

\textbf{Task 3: Mango}

\begin{tcolorbox}[onecol, adjusted title=Raw images]

User: There are two mangoes in the image, one on the left and another on the right. Step by step, touch and identify the properties of the mangoes, and pick up the riper one.

\vspace{2mm}

Robot: Touch the mango on the left to assess its softness.

\vspace{2mm}

User: The first image is the gelsight tactile sensor image before touching the object: <image1/>. \\The second image is the gelsight tactile sensor image after touching the object:  <image2/>

\vspace{2mm}

Robot: The mango on the left is firm. Now, touch the mango on the right to assess its softness.

\vspace{2mm}

User: The first image is the gelsight tactile sensor image before touching the object: <image1/>. \\The second image is the gelsight tactile sensor image after touching the object:  <image2/>

\vspace{2mm}

Robot: The mango on the right is also firm. Since both mangoes are firm, pick up either one.

\end{tcolorbox}

\begin{tcolorbox}[onecol, adjusted title=Ours]

User: There are two mangoes in the image, one on the left and another on the right. Step by step, touch and identify the properties of the mangoes, and pick up the riper one.

\vspace{2mm}

Robot: Touch the mango on the left to assess its softness. Please provide feedback on the softness.

\vspace{2mm}

User: The hardness level is: 4.293282985687256, The roughness level is: 5.726420879364014

\vspace{2mm}

Robot: Touch the mango on the right to assess its softness. Please provide feedback on the softness.

\vspace{2mm}

User: The hardness level is: 3.2263505458831787, The roughness level is: 5.662842750549316
\vspace{2mm}
Robot: Pick up the mango on the right, as it is softer and likely riper.

\end{tcolorbox}

\section{Ablation on Tactile Feedback Format from Octopi}
\label{app:ablation}

In the original Octopi framework, the tactile encoder is first trained alongside a classifier to predict physical properties on a continuous scale from 1 to 10. In the second phase, the classifier is removed, and a language model is trained to align with the encoder's embeddings. To facilitate alignment with language, the physical properties are discretized into categories (e.g., three categories for roughness: smooth, slightly rough, and rough). However, in our case, we observed that the discretized descriptions failed to distinguish between objects with subtle differences due to their limited granularity. In our ablation experiments, we tested the effect of (a) using the full Octopi pipeline, with its language model generating a complete language description of the object, versus (b) directly using the continuous output from the classifier. Results show that in the wiping task, both methods achieved a 100\% success rate, as the roughness levels of the two sponges differed significantly. However, in the mango peeling task, where the difference in hardness is more subtle, method (a) achieved only a 60\% success rate, while method (b) reached 75\%. This highlights the limitations of using natural language to represent physical properties sensed through touch.


\section{Interpolant Controller}
\label{app:interpolant}

\textbf{Interpolant Model.}
Bridger~\cite{chen2024behavioral} leverages stochastic interpolant to bridge arbitrary source and target action distributions. Unlike conventional diffusion-based imitation learning methods that denoise from standard Gaussian noise, Bridger can utilize more informative source policies as starting points, which leads to better performance with fewer diffusion steps.
This is particularly valuable in our context, where the VLA policy already captures many aspects of the target behavior but lacks the refinement that tactile sensing provides.

Mathematically, Bridger employs a stochastic interpolant that defines a continuous-time stochastic process between source policy $\pi_0$ and target policy $\pi_1$ with:

\begin{equation}
a_t = I(t, a_0, a_1, x) + \gamma(t)z
\end{equation}

\noindent where $I$ is an interpolant function with boundary conditions $I(0, a_0, a_1, x) = a_0$ and $I(1, a_0, a_1, x) = a_1$, $\gamma(t)$ controls the noise schedule with $\gamma(0) = \gamma(1) = 0$, and $z$ is standard Gaussian noise.

The forward stochastic differential equation that guides our refinement process is:

\begin{equation}
da_t = b_F(t, a_t, x)dt + \sqrt{2\epsilon(t)}dW_t
\end{equation}

\noindent where $b_F$ is a velocity function, $\epsilon(t)$ controls diffusion strength, and $W_t$ is a Wiener process. 

This formulation allows us to incorporate tactile sensing into the refinement process, ensuring that the resulting policy benefits from both the robustness of the VLA model and the precise contact-rich manipulation capabilities enabled by tactile feedback.

\textbf{Interpolant Controller.}
We adopt the interpolant model to refine the VLA policy with tactile sensing. Our approach uses the VLA model as the source policy $\pi_{VLA}$, which provides a foundation for manipulation tasks based on visual and proprioceptive feedback. The target policy$\pi_{Exp}$ is the expert policy that incorporates tactile feedback for contact-rich manipulation.


\textbf{Data Collection.} For the three tasks evaluated, we collected datasets through kinesthetic teaching: \textbf{Cup}: Pick (40 episodes), Place (60 episodes), in totoal 8 mins data; \textbf{Wipe}: Pick (40 episodes), Wipe (60 episodes), in totoal 15 mins data; \textbf{Peel}: Pick (60 episodes), Peel (120 episodes), in total 30 mins data. Sensor information was collected at 10 Hz, including 2 camera images, Gelsight images and robot proprioceptive (end-effector pose and gripper state). We labeled each episode with a corresponding linguistic instruction to facilitate instruction-following for VLA.

We use a single RTX 4090 GPU for finetuning.
We first finetuned RDT on unrelated Franka data (100k steps, $\approx$80 hrs) to align action spaces.
Next, we finetuned the pretrained VLA model on the collected dataset without tactile modality. Each task required 20k training steps ($\approx$16 hrs). Then, we used this finetuned model to predict action chunks conditioned on the observations in our dataset. By incorporating tactile input, we constructed a dataset $D_{\text{VLA}}$ for interpolant controller learning. For each episode with length $T$, we constructed: $\{s_{t-n:t}, O^p_{t-n:t}, a_{t:t+T_a}, a^{\text{Exp}}_{t:t+T_a}\}_{t=n}^{T}$ for training, where $a^{Exp}$ refers to actions from expert demonstrations, $n$ refers to the horizon of history condition, $T_a$ refers to action chunk horizon. We adopted $T_a = 64, n = 2$ from the VLA pretraining settings for our experiments.

\textbf{Controller Training.}
We feed tactile images $ o^m_t$ into the marker tracking algorithm based on the OpenCV library to obtain a low dimensional tactile state $m_t$ (force vectors). The RGB images in $s_t$ are fed into a pretrained DinoV2 model to obtain visual embeddings $z^o_t$. This is concatenated with the robot proprioceptive state $z^r_t$ to form the interpolant controller input observation: $z_t := [z^r_{t-n:t},z^o_{t-n:t},m_{t-n:t}]$.

The Interpolant controller $\pi_I (\hat a_{t:t+T_a}|a_{t:t+T_a}, z_t)$ is trained to generate refined action chunk $\hat a_{t:t+T_a}$ to minimize the MSE loss with respect to expert actions:
\begin{equation}
    \min_{\theta} \frac{1}{|D_{\text{VLA}}|}\sum || a^{\text{Exp}}_{t:t+T_a} - \pi_{I(\hat a_{t:t+T_a}|a_{t:t+T_a}, z_t;\theta)}||^2
\end{equation}

for $(a_{t:t+T_a}, z_t, a_{t:t+T_a}^{\text{Exp}})$ in  $D_{VLA}$.

\textbf{Controller Inference.}
During inference, the first 48 steps of the 64-step action chunk generated by the VLA model are used for action refinement. The VLA model updates observations and generates new action chunks after the current truncated action chunk refined and executed. The interpolant controller updates observations and refines action chunks at most 8 Hz, with the refined actions executed at 8 Hz by a Cartesian controller with simple PD control, an exception is for task Mango Peeling, we use a impedance controller for execution.


\immediate\write18{cp \jobname.bbl main.bbl}
\end{document}